\documentclass[11pt]{article}

\usepackage[utf8]{inputenc}
\usepackage[T1]{fontenc}
\usepackage{times}
\usepackage[margin=1in]{geometry}
\usepackage{graphicx}
\usepackage{amsmath,amssymb}
\usepackage{booktabs}
\usepackage{multirow}
\usepackage{url}
\usepackage{hyperref}
\usepackage{xcolor}
\usepackage{natbib}
\usepackage{enumitem}
\usepackage{subcaption}

\graphicspath{{./figures/}}

\hypersetup{
    colorlinks=true,
    linkcolor=blue,
    citecolor=blue,
    urlcolor=blue
}

\title{LLM-Metrics: Measuring Research Impact Through\\Large Language Model Memory}

\author{
  Si Shen \\
  School of Economics and Management \\
  Department of Computer Science and Engineering \\
  Nanjing University of Science and Technology \\
  Nanjing, 210094, China
  \and
  Wenhua Zhao \\
  College of Information Management \\
  Nanjing Agricultural University \\
  Nanjing, 210095, China
  \and
  Danhao Zhu\thanks{Corresponding author.} \\
  Department of Criminal Science and Technology \\
  Jiangsu Police Institute \\
  Nanjing, 210031, China \\
  \texttt{ZHUDANHAO@JSPI.EDU.CN}
}

\date{}

\begin{document}
\maketitle

\begin{abstract}
Citation counts remain the dominant metric for assessing research impact, yet they suffer from well-documented limitations: temporal lag, disciplinary bias, and Matthew effects. Here we propose LLM-Metrics, a research-impact assessment metric derived from the parametric memory of large language models (LLMs). The central hypothesis is that high-impact papers receive greater exposure in the academic community, that this exposure enters LLM training data in textual form, and that models consequently form stronger parametric memory of these papers. We designed four types of multiple-choice probes (title recognition, author recognition, method recognition, and venue recognition) and evaluated 549 computer science papers published in 2023--2024 across 17 LLMs spanning 0.5B to 72B parameters from six vendors. Of the 17 models, 15 produced positive predictions (9 significant at $p < 0.05$), with an overall Spearman correlation of $\rho = 0.1495$ ($p = 0.0004$) against citation counts. Three additional findings support the proposed mechanism. First, the predictive signal was stronger for 2024 papers ($\rho = 0.1880$), whose citation counts were near zero at model-training time, ruling out a simple reverse-causality explanation. Second, author-recognition probes showed the strongest discriminative power, consistent with an exposure-driven memory mechanism. Third, model scale and predictive power were non-monotonic: a 3B-parameter model (Llama-3.2-3B-Instruct, $\rho = 0.1829$) outperformed most larger models, supporting a selective-memory hypothesis in which the limited capacity of smaller models can serve as an effective information filter. LLM-Metrics offers a real-time, cross-disciplinary, citation-independent paradigm for research assessment.
\end{abstract}

\section{Introduction}
\label{sec:intro}

A central task in scientometrics is measuring the impact of scientific work. Since Garfield introduced the Science Citation Index~\citep{garfield1955citation}, citation counts have served as the dominant indicator of research influence. However, three widely acknowledged limitations constrain their utility. First, citations accumulate slowly: a paper typically requires three to five years to gather enough citations for meaningful assessment~\citep{natbiomedeng2022}, rendering citation counts nearly useless for rapidly emerging fields. Second, citation practices vary substantially across disciplines: mathematics papers receive far fewer citations on average than biomedical papers, making cross-disciplinary comparisons unreliable~\citep{costas2015altmetrics, gonzalez2025altmetrics}. Third, the Matthew effect amplifies the visibility of established authors and prestigious institutions, so that citation counts partly reflect academic reputation rather than the intrinsic quality of the work~\citep{merton1968matthew, soliman2025precocious}.

Alternative metrics (altmetrics) have sought to address these shortcomings by incorporating social media mentions, news coverage, and policy citations~\citep{priem2010altmetrics}. Yet altmetrics primarily capture societal attention rather than scholarly impact, and they inherit many of the same disciplinary biases and temporal constraints as traditional citations.

The emergence of large language models (LLMs) opens a new paradigm for scientometrics. Modern LLMs are trained on vast corpora that include substantial academic content: arXiv preprints, conference proceedings, technical blogs, and scholarly discussions~\citep{vaswani2017attention, brown2020language}. Recent work has demonstrated that LLMs can be leveraged for academic tasks ranging from conference organization to scientific reasoning~\citep{luo2025leveraging, li2025self}. The parametric memory that models form of this content is not uniform. Papers that appear frequently in training corpora can become more firmly encoded in model parameters, whereas long-tail papers may be partially encoded or forgotten. We hypothesize that this differential memory pattern constitutes a measurable trace of scholarly exposure, because papers that attract discussion, sharing, replication, and citation are more likely to recur across academic text sources.

Building on this observation, we propose \textbf{LLM-Metrics}, a paper-level research-impact assessment metric grounded in LLM memory scores. The core hypothesis is that \textbf{the differential memory that LLMs form of academic papers provides a real-time signal of scholarly exposure and later citation impact}. This hypothesis rests on a four-step logic chain: (1) high-impact papers tend to receive greater exposure in the academic community; (2) this exposure enters LLM training data as text; (3) higher training-data exposure leads to stronger parametric memory; and (4) carefully designed probes can estimate paper-level memory strength. LLM-Metrics therefore turns model memory into a measurable scientometric signal rather than treating LLMs only as prediction tools.

The contributions of this work are as follows:
\begin{enumerate}[nosep]
    \item We define LLM-Metrics, a quantitative research-impact metric based on paper-level model memory, and provide a reproducible computational pipeline.
    \item We conduct a large-scale empirical evaluation across 17 LLMs (six vendors, six model families, 0.5B--72B parameters) on 549 papers, showing that model memory contains a statistically reliable signal of later citation impact.
    \item We validate the selective-memory hypothesis through multiple converging lines of evidence: a temporal split test, probe-type differentiation, and non-monotonic scaling with model size.
    \item We characterize vendor- and family-level differences in predictive performance, showing that model choice is a substantive component of memory-based research assessment.
\end{enumerate}

\clearpage
\begin{figure}[t!]
\centering
\includegraphics[width=\textwidth]{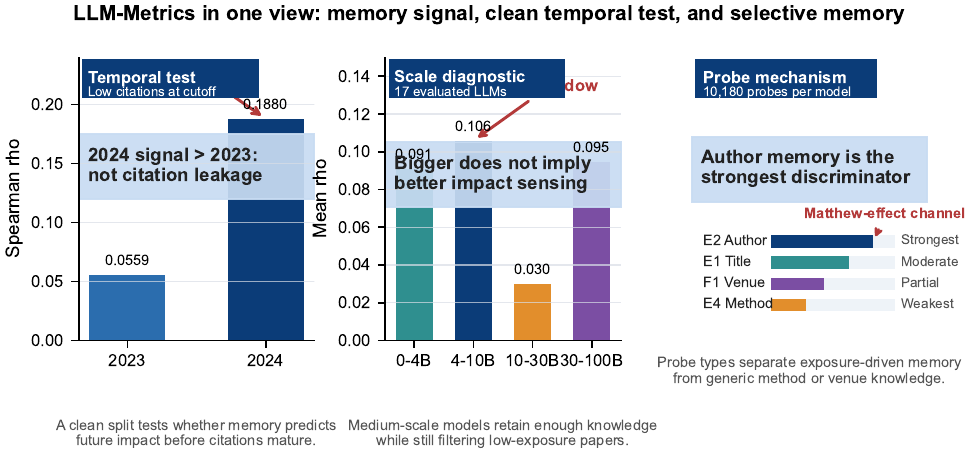}
\caption{\textbf{Front summary of LLM-Metrics.} This overview condenses the paper's three central empirical claims using only results reported in this study. \textbf{Left}, the 2024 cohort shows a stronger overall correlation with later citations than the 2023 cohort ($\rho=0.1880$ vs. $\rho=0.0559$), supporting a citation-independent exposure signal. \textbf{Middle}, predictive power is non-monotonic with model size: the 4--10B group has the highest mean $\rho$, while the 10--30B group is weakest. \textbf{Right}, probe-type evidence identifies author recognition as the strongest discriminative channel, consistent with exposure-driven selective memory.}
\label{fig:front_overview}
\end{figure}

\section{Related Work}
\label{sec:related}

\subsection{Academic Impact Assessment and Citation Prediction}

Traditional bibliometric approaches rely on citation-based metrics, including raw citation counts, journal impact factors~\citep{garfield2006history, garfield1955citation}, and the h-index~\citep{hirsch2005index}. While widely adopted, these metrics are inherently retrospective and subject to well-documented biases: field-dependent citation norms, self-citation inflation, and Matthew effects that amplify the visibility of already-prominent authors and institutions~\citep{merton1968matthew, costas2015altmetrics, gonzalez2025altmetrics}. Prior critiques of academic metrics show how metric-driven evaluation can distort research behavior~\citep{fire2019over}. Recent reviews further emphasize ethical concerns including gender bias, geographical bias, and metric gaming~\citep{ghousi2025ethics}.

Alternative approaches have emerged to address these limitations. Altmetrics~\citep{priem2010altmetrics} capture online attention through social media mentions, news coverage, policy citations, and reference manager saves. Prior studies report moderate correlations between altmetric scores and citation counts across disciplines~\citep{thelwall2013altmetrics} and show that altmetric indicators capture dimensions of attention that are partly orthogonal to traditional citations~\citep{costas2015altmetrics}. Other work develops evaluation frameworks and reviews the strengths and limitations of altmetrics~\citep{arroyo2023evaluative, gonzalez2025altmetrics}, while domain-specific studies analyze social-media visibility and second-order citation pathways~\citep{gholampour2024social, alperin2024second}. However, altmetrics primarily capture societal attention rather than scholarly impact, and they inherit many of the same disciplinary biases as traditional metrics.

In the domain of citation prediction, forecasting future citation counts has been an active research area. Existing methods use author reputation, venue prestige, early citation patterns, text embeddings, main-text structure, temporal citation dynamics, collaboration networks, semantic metadata, and evolving knowledge graphs~\citep{zhao2020citation, vital2024predicting, hirako2024cimate, abrishami2019forecasting, ma2021citation, zhu2024instant, gu2025forecasting}. Recent systems such as ForeCite adapt pre-trained language models for citation-rate regression at large scale~\citep{hull2025forecite}, and prompt-based LLM methods can predict highly cited papers using information available at publication time~\citep{ye2026llmpredict}. These methods are strong predictors, but they primarily model which observable features are associated with later citations. LLM-Metrics asks a different question: whether the parametric memory of LLMs can itself serve as a research-impact metric. This framing shifts LLMs from supervised predictors to measurement instruments for academic exposure.

\subsection{LLM Knowledge Representation and Memorization}

Prior work shows that pre-trained language models encode substantial factual knowledge without fine-tuning~\citep{petroni2019language}, that larger models tend to perform better on closed-book question answering~\citep{roberts2020knowledge}, and that factual recall depends strongly on training-data frequency~\citep{kandpal2023large}. These findings motivate our selective-memory hypothesis. Subsequent work has explored knowledge editing~\citep{meng2022locating, meng2023mass}, knowledge distillation~\citep{gu2024minillm}, multifaceted knowledge recall benchmarks~\citep{zhao2024belief}, and large-scale belief extraction from frontier models~\citep{ghosh2025mining}.

Training-data memorization has been studied extensively from privacy and security perspectives. LLMs can reproduce verbatim training sequences~\citep{carlini2021extracting}, and formal measurement frameworks quantify memorization across neural language models~\citep{carlini2023quantifying}. Analyses using model suites and controlled settings show that memorization varies with scale, data repetition, training dynamics, deduplication, and prompting strategy~\citep{biderman2023pythia, xiong2025landscape, chang2024factual, bordt2024elephants, kiyomaru2024comprehensive, borec2024unreasonable}. Closest to our setting, citation frequency has been linked to the factual accuracy of LLM-generated references~\citep{niimi2025hallucinate}, suggesting that bibliographic facts about prominent papers are more likely to be retained.

Collectively, these studies motivate a core premise: LLMs do not uniformly memorize entities in their training data; rather, memorization strength is often tied to exposure frequency. Our methodology tests whether this premise extends to academic papers and whether paper-level memory strength is predictive of later citation impact.

\subsection{Knowledge Probing and Scientific Evaluation}

Knowledge probes have recently been used to estimate LLM parameter counts, evaluate scientific reasoning, and assess scientific understanding~\citep{li2026ikp, guo2026scivqr, feng2024sciknoweval}. LLMs have also been applied to academic workflows and scientific response generation~\citep{luo2025leveraging, li2025self}. For impact prediction, fine-tuned LLMs can predict article impact from titles and abstracts~\citep{wang2024words}, but this line of work still treats the LLM as a supervised predictor rather than using parametric memory as the signal being measured.

Despite these advances, a gap remains: no existing work has systematically investigated whether the \textit{differential memorization} of academic papers across LLMs can serve as a scientometric metric. This gap is significant because LLMs are increasingly trained on academic corpora, while citation-based metrics remain slow and field-dependent. Our work addresses this gap by introducing LLM-Metrics, which repurposes memory probes---traditionally used for model diagnostics---as instruments for measuring paper-level exposure traces. The key insight is that the degree to which a paper is ``remembered'' by independently trained LLMs reflects how strongly that paper is represented across academic text ecosystems. This approach extends knowledge probing from model analysis to real-time scientometric measurement.

\section{Methods}
\label{sec:method}

\subsection{Definition of LLM-Metrics}

LLM-Metrics is a paper-level research-impact metric derived from LLM probe scores. It measures how strongly a model remembers a paper and tests whether that memory strength predicts later citation impact without using citation counts as input. Its computation proceeds in three steps, each designed to balance theoretical grounding with computational feasibility.

\textbf{Step 1: Probe scoring.} For each paper $p$, we design $K$ probes (multiple-choice questions) and present them to model $m$. Each probe response is classified into one of five grades (Table~\ref{tab:grades}). The five-grade scheme was motivated by several considerations. First, distinguishing ``correct'' from ``incorrect'' is the minimal requirement, but a binary classification loses information---when a model attempts to answer but does not fully succeed (e.g., it eliminates some distractors but fails to identify the correct one), its behavior is qualitatively different from random guessing, hence the introduction of PARTIAL (+1). Second, REFUSAL (0 points) provides a unique signal: a model choosing not to answer rather than guessing is closely related to its safety alignment and knowledge-boundary awareness. Third, HALLUCINATION ($-2$), as the most severe error type---the model fabricates non-existent information---carries the heaviest penalty, as hallucinations are more dangerous than simple errors and can mislead academic assessment.

\begin{table}[htbp]
\centering
\caption{Five-grade scoring scheme for probe responses.}
\label{tab:grades}
\begin{tabular}{lcl}
\toprule
\textbf{Grade} & \textbf{Score} & \textbf{Meaning} \\
\midrule
CORRECT & +2 & Model selects the correct answer \\
PARTIAL & +1 & Model eliminates some distractors but does not reach the correct answer \\
REFUSAL & 0 & Model explicitly declines to answer (``I don't know'') \\
WRONG & $-1$ & Model selects an incorrect answer \\
HALLUCINATION & $-2$ & Model fabricates non-existent information \\
\bottomrule
\end{tabular}
\end{table}

\textbf{Step 2: Score mapping.} While the five-grade scheme captures rich memory signals, its multi-level discreteness can introduce unnecessary granularity differences during aggregation (e.g., different models may exhibit different score distributions across probe types). To simplify interpretation and ensure cross-model comparability, we map the five grades to the $[0, 1]$ interval to produce a binarized memory signal:
\begin{equation}
    \text{score}_{p,m,k}^{\text{bin}} = \begin{cases}
        1 & \text{if CORRECT (+2)} \\
        0.5 & \text{if PARTIAL (+1)} \\
        0 & \text{if REFUSAL / WRONG / HALLUCINATION}
    \end{cases}
\end{equation}

This mapping assigns CORRECT to 1 (``remembered''), PARTIAL to 0.5 (``partially remembered''), and the remainder to 0 (``not remembered''). Notably, REFUSAL is classified as ``not remembered'' rather than as a separate category---this design choice reflects a key assumption: a model's behavior when facing knowledge gaps (refusing vs. producing errors) reflects its output strategy more than memory strength per se.

\textbf{Step 3: Aggregation.} The binarized scores are averaged across all $K_{p,m}$ probes for paper $p$ on model $m$:
\begin{equation}
    \text{LLM-Metrics}_m(p) = \frac{1}{K_{p,m}} \sum_{k=1}^{K_{p,m}} \text{score}_{p,m,k}^{\text{bin}}
\end{equation}

LLM-Metrics$_m(p)$ takes values in $[0, 1]$, with higher values indicating stronger model $m$ memory of paper $p$. When all $K$ probes are answered correctly, LLM-Metrics $= 1$ (perfect memory); when none are answered correctly, LLM-Metrics $= 0$ (no memory). In practice, LLM-Metrics values average around 0.35--0.40, reflecting moderate overall memory levels for academic papers.

The core theoretical assumption of LLM-Metrics is that a paper's memory strength across multiple independently trained models reflects its exposure within academic text ecosystems. By computing LLM-Metrics independently across models that differ in vendor, scale, and training cutoff, we test whether the memory signal is cross-model consistent. If independently trained models produce similar memory rankings for the same set of papers, LLM-Metrics captures a shared exposure signal rather than an idiosyncrasy of any single model's training data.

\subsection{Probe Design}

We designed four types of multiple-choice probes, yielding 10,180 probes per model, covering distinct dimensions of paper memory:

\begin{table}[htbp]
\centering
\caption{Probe types and their per-model counts.}
\label{tab:probes}
\begin{tabular}{lcc}
\toprule
\textbf{Probe Type} & \textbf{Per-Model Count} & \textbf{Capability Assessed} \\
\midrule
E1\_mc (Title Recognition) & 2,745 & Whether the model remembers the paper title \\
E2\_mc (Author Recognition) & 2,745 & Whether the model remembers the paper authors \\
E4\_mc (Method-Paper) & 1,945 & Whether the model remembers the paper's method \\
F1\_mc (Venue Information) & 2,745 & Whether the model remembers the publication venue and year \\
\bottomrule
\end{tabular}
\end{table}

Each probe contains one correct option and three to four distractors. Distractors were constructed according to three strict principles: (1) they belong to the same semantic category as the correct option (e.g., E1 distractors are titles of other real papers in the same field, E2 distractors are names of other real authors in the same field), ensuring that models cannot eliminate incorrect options based solely on domain or topic information; (2) they are superficially plausible, preventing models from eliminating them through simple lexical or format matching---all options maintain consistent phrasing style, length, and format; (3) the correct answer cannot be inferred from context and must rely on parametric knowledge stored in model parameters---the probe question itself provides no contextual clues that would help infer the correct answer.

To illustrate, an E1\_mc (Title Recognition) probe is formatted as: ``Which of the following is the title of the paper [Paper Title]?'', with the correct option being the actual title and distractors being other paper titles randomly sampled from the same field. An E2\_mc (Author Recognition) probe is formatted as ``Who are the authors of the paper [Paper Title]?'', with the correct option being the actual authors and distractors being other prominent authors in the field. The E4\_mc (Method-Paper) probe type contains fewer items (1,945 vs. 2,745) because not all papers introduce a distinct, nameable method---for instance, survey papers or theoretical papers may not propose specifically named methods.

\begin{figure}[htbp]
\centering
\includegraphics[width=\textwidth]{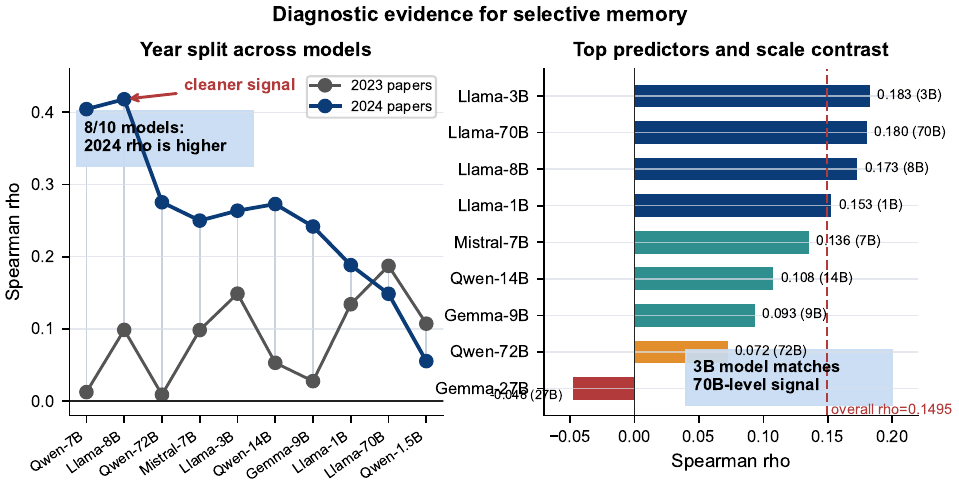}
\caption{\textbf{Diagnostic evidence for selective memory.} \textbf{Left}, model-level year-split analysis shows that 8 of 10 models achieve higher Spearman $\rho$ on 2024 papers than on 2023 papers, even though 2024 papers had minimal citation accumulation at training time. \textbf{Right}, top model ranking and scale contrast show that Llama-3.2-3B reaches the highest correlation ($\rho=0.1829$), matching the 70B model-level signal and exceeding most larger models; the dashed line marks the overall ensemble $\rho=0.1495$.}
\label{fig:front_diagnostics}
\end{figure}

Overall, the core principle underlying our probe design strategy is to estimate how strongly a model remembers multiple metadata dimensions of a paper. We then test whether this estimated memory strength predicts later citation impact. This strategy avoids using citation counts as model inputs and instead treats the LLM as a real-time instrument for measuring traces of academic dissemination.

\subsection{Paper Sample}

We selected 549 computer science papers published in 2023--2024, spanning artificial intelligence, systems, theory, security, and other subfields. Summary statistics of the sample are shown in Table~\ref{tab:paper_stats}.

\begin{table}[htbp]
\centering
\caption{Summary statistics of the paper sample.}
\label{tab:paper_stats}
\begin{tabular}{lc}
\toprule
\textbf{Metric} & \textbf{Value} \\
\midrule
Total papers & 549 \\
2023 papers & 415 (75.6\%) \\
2024 papers & 134 (24.4\%) \\
Top-venue papers & 31 (5.6\%) \\
Mean citations & 74.6 \\
Median citations & 24 \\
Citation range & 0--1,674 \\
\bottomrule
\end{tabular}
\end{table}

Citation data were obtained from the Semantic Scholar API in May 2026, approximately 1.5--2.5 years after publication, sufficient for meaningful ground-truth signals to accumulate. Our choice of the 2023--2024 window carries a specific research design rationale: at the time of model training cutoff (mid-to-late 2024, see Table~\ref{tab:models}), 2023 papers had at most $\sim$1.5 years of citation accumulation, while 2024 papers had at most $\sim$6 months, with most 2024 papers having only 0--2 citations. This temporal disparity creates a natural contrast: 2024 papers had near-zero citation signal at training time, so if LLM-Metrics is still associated with their subsequent citations, the result rules out a simple memorized-citation-count explanation (detailed analysis in \S4.3).

The paper sample was constructed to ensure diversity across multiple dimensions. Papers were drawn from a broad range of venues, including top-tier conferences (CCF-A), mid-tier conferences (CCF-B, CCF-C), other ranked venues, and unranked venues/preprints (primarily arXiv). The sample spans a wide range of citation counts (0--1,674), with a heavy right-skewed distribution typical of bibliometric data: the median (24) is substantially lower than the mean (74.6), reflecting the concentration of citations among a small number of highly influential papers. Approximately 20\% of the most-cited papers account for over 60\% of total citations. This extreme skew has important implications for our choice of statistical methods: it violates the normality assumption required for Pearson correlation, motivating our primary reliance on Spearman's rank correlation.

For the paper selection and sampling strategy, we employed a combination of stratified random sampling and targeted inclusion. We first retrieved metadata for all searchable computer science papers from 2023--2024 via the Semantic Scholar API and stratified them by citation count (0, 1--9, 10--49, 50--99, 100--249, 250--499, 500+), ensuring adequate representation across all citation tiers. We also purposively included 31 papers from CCF-A top-tier venues to cover representative high-impact samples, along with a number of arXiv preprints to cover papers not yet formally published. This sampling strategy ensures sufficient variation along the citation dimension, providing the foundation for the quantile-based interval analyses in \S4.4.

\subsection{Model Sample}

We evaluated 17 LLMs spanning six vendors, six model families, and parameter counts from 0.5B to 72B, as shown in Table~\ref{tab:models}. This selection was designed to maximize vendor diversity, scale variation, and training cutoff spread, enabling a broad stress test of LLM-Metrics across heterogeneous models.

\begin{table}[htbp]
\centering
\caption{LLMs evaluated in this study. All models were served through vLLM~\citep{kwon2023vllm} at full precision.}
\label{tab:models}
\begin{tabular}{lcccc}
\toprule
\textbf{Model} & \textbf{Params} & \textbf{Vendor} & \textbf{Family} & \textbf{Cutoff} \\
\midrule
Qwen2.5-0.5B-Instruct & 0.5B & Alibaba & Qwen2.5 & 2024-09 \\
Qwen2.5-1.5B-Instruct & 1.5B & Alibaba & Qwen2.5 & 2024-09 \\
Qwen2.5-3B-Instruct & 3.0B & Alibaba & Qwen2.5 & 2024-09 \\
Qwen2.5-7B-Instruct & 7.0B & Alibaba & Qwen2.5 & 2024-09 \\
Qwen2.5-14B-Instruct & 14.0B & Alibaba & Qwen2.5 & 2024-09 \\
Qwen2.5-32B-Instruct & 32.0B & Alibaba & Qwen2.5 & 2024-09 \\
Qwen2.5-72B-Instruct & 72.0B & Alibaba & Qwen2.5 & 2024-09 \\
gemma-2-2b-it & 2.0B & Google & Gemma-2 & 2024-06 \\
gemma-2-9b-it & 9.0B & Google & Gemma-2 & 2024-06 \\
Gemma-2-27B & 27.0B & Google & Gemma-2 & 2024-06 \\
Llama-3.2-1B-Instruct & 1.0B & Meta & LLaMA-3 & 2024-09 \\
Llama-3.2-3B-Instruct & 3.0B & Meta & LLaMA-3 & 2024-09 \\
Llama-3.1-8B-Instruct & 8.0B & Meta & LLaMA-3 & 2024-07 \\
Llama-3.3-70B-Instruct & 70.0B & Meta & LLaMA-3 & 2024-12 \\
Phi-4-mini-instruct & 4.0B & Microsoft & Phi & 2024-12 \\
Mistral-7B-Instruct-v0.3 & 7.0B & Mistral AI & Mistral & 2024-05 \\
glm-4-9b-chat & 9.0B & Zhipu AI & GLM & 2024-06 \\
\bottomrule
\end{tabular}
\end{table}

Model selection followed three core criteria. First, vendor diversity: covering the principal LLM developers---Alibaba (7 models), Google (3), Meta (4), Microsoft (1), Mistral AI (1), and Zhipu AI (1). Different vendors employ distinct training data compositions, data preprocessing pipelines, architectural designs, and instruction-tuning strategies. If LLM-Metrics produces positive predictions across all or most vendors, it indicates that the memory signal is robust across different training data distributions.

Second, a dense scale gradient: parameter counts span over two orders of magnitude, from 0.5B (Qwen2.5-0.5B-Instruct) to 72B (Qwen2.5-72B-Instruct), with fine-grained intermediate sizes---0.5B, 1.0B, 1.5B, 2.0B, 3.0B, 4.0B, 7.0B, 8.0B, 9.0B, 14.0B, 27.0B, 32.0B, 70.0B, 72.0B. This dense gradient enables a rigorous test of whether the relationship between model scale and LLM-Metrics predictive power exhibits non-monotonic patterns (\S4.5), a unique prediction of the selective memory hypothesis.

Third, training cutoff control: all models have training cutoffs between May and December 2024. This temporal window has particular research design value: models may have encountered 2023--2024 papers during training (e.g., arXiv preprints, PDF full texts) but could not have observed mature subsequent citation counts, which had not yet accumulated at training time. Therefore, LLM-Metrics' predictive power cannot be explained by models simply ``learning'' mature citation information. All models were used at full precision; quantized variants were excluded to avoid precision loss.

\subsection{Evaluation Protocol}

All models were served through the vLLM framework~\citep{kwon2023vllm} and evaluated in parallel across four NVIDIA GPUs. vLLM provides efficient memory management and continuous batching capabilities, enabling us to complete the large-scale evaluation of 17 models $\times$ 10,180 probes (approximately 173,000 inference calls) within a reasonable time window. Each model was deployed independently, avoiding interference from loading and unloading different model weights, ensuring that every model ran in a clean GPU memory environment.

Temperature was set to zero to ensure deterministic output. In typical conversational applications, a non-zero temperature is used to increase response diversity and creativity; however, in knowledge probing tasks, diversity is not the goal---our objective is to measure the deterministic knowledge stored in model parameters, not the model's stochastic sampling behavior at a given temperature. Setting temperature to zero ensures that each query for the same probe yields exactly the same output, guaranteeing reproducibility of evaluation results.

Model responses were parsed via regular expressions into the five-grade scoring scheme (CORRECT, PARTIAL, REFUSAL, WRONG, HALLUCINATION). Parsing rules included: identifying whether the model explicitly pointed to an option (e.g., ``A'', ``option A'', ``the answer is A''), recognizing refusal patterns (e.g., ``I don't know'', ``cannot determine''), and detecting nonsensical or completely irrelevant output (HALLUCINATION). Responses whose format did not match expectations (e.g., non-option text not belonging to any refusal pattern) were marked as ERROR and excluded from subsequent analysis. In practice, ERROR rates were extremely low ($<0.5\%$), indicating that most models can largely follow the multiple-choice instruction format.

The entire evaluation pipeline was orchestrated by automated Python scripts. After each model's evaluation was complete, raw responses (including full model output text and parsed scores) were stored as structured JSON files, from which subsequent analyses read data. This pipeline design ensures traceability and reproducibility of data processing.

\subsection{Statistical Analysis}

This study employs multiple complementary statistical methods to evaluate the validity, consistency, and limitations of LLM-Metrics.

\textbf{Primary correlation analysis.} The correlation between LLM-Metrics and citation counts is assessed using Spearman's rank correlation coefficient ($\rho$) for three reasons: (1) citation counts are highly skewed (median 24, mean 74.6), violating the normality assumption required for Pearson correlation, while Spearman's coefficient makes no distributional assumptions; (2) Spearman correlation is sensitive to monotonic relationships without requiring linearity, consistent with our theoretical expectation of monotonically increasing memory strength with citation exposure; (3) Spearman correlation is less sensitive to outliers, which is particularly important in the heavy-tailed citation distribution---a few extremely highly cited papers will not disproportionately dominate the correlation estimate. Spearman $\rho$ is computed independently for each model, correlated against the paper's total citation count.

The reported $p$-values are two-sided $p$-values from the Spearman rank correlation test, testing the null hypothesis $H_0: \rho = 0$ (no monotonic relationship between memory scores and citation counts). Results with $p < 0.05$ are considered statistically significant, $p < 0.01$ highly significant, and $p < 0.001$ extremely significant. Since we conduct tests on 17 models separately, multiple comparisons are a concern. We choose not to apply strict Bonferroni correction (corrected threshold $p < 0.0029$) for two reasons: (1) our core interest lies not in whether individual models are ``significant'' but in the overall directional consistency and effect-size pattern across models; (2) in exploratory research, overly strict correction increases Type II error risk. Instead, we assess the non-randomness of cross-model direction via a binomial sign-consistency test.

\textbf{Ensemble LLM-Metrics.} To obtain a model-ensemble overall LLM-Metrics, we take the median (rather than the mean) of scores across all 17 models. The rationale for choosing the median is that memory scores can exhibit systematic shifts across models due to differing output strategies (active vs. cautious), making the median more robust to extremes than the mean. For example, a cautious model that refuses 97\% of questions has a fundamentally different score distribution than an active model that almost never refuses; using the mean would overemphasize the active model's contribution while neglecting the cautious model's signal. The median treats each model's ranking information equally.

\textbf{High-low difference analysis.} For the high-low difference analysis, papers are divided into a high-citation group (top 25\%) and a low-citation group (bottom 25\%). This design follows the classic ``extreme-group comparison'' paradigm in scientometrics: if LLM-Metrics contains citation-related information, the paper groups corresponding to opposite ends of the citation spectrum should exhibit the largest memory-score differences. The mean CORRECT rate is computed for each group, and the high-low difference is defined as the high-group CORRECT rate minus the low-group CORRECT rate (in percentage points). Between-group differences are assessed using independent-samples $t$-tests. The middle 50\% of papers are excluded to maximize between-group contrast.

\textbf{Cross-model consistency analysis.} We employ two complementary methods for assessing cross-model consistency. First, sign-consistency rate: we compute the proportion of models producing positive $\rho$ and compare it against the random expectation of 50\% via a binomial test. This metric focuses only on prediction direction (positive/negative), ignoring effect size, providing the most conservative consistency assessment. Second, pairwise correlation: we compute the Spearman correlation coefficient of LLM-Metrics scores between every pair of models, as a measure of inter-model memory pattern similarity. High pairwise correlation ($r > 0.5$) indicates that two models' memory rankings of the same paper set are highly consistent; low pairwise correlation ($r < 0.2$) suggests significant divergence in memory patterns. The pairwise correlation matrix is further analyzed by vendor grouping (within-vendor vs. cross-vendor) to examine the influence of shared training data on memory patterns.

\textbf{Citation interval analysis.} The citation interval analysis groups papers by citation count and computes each group's mean memory score, CORRECT rate, WRONG rate, and REFUSAL rate. Interval partitioning employs a hybrid strategy: fine-grained intervals in the low-citation range (0, 1--4, 5--9) to capture subtle memory-score variations, and coarse-grained intervals in the high-citation range (50--99, 100--199, 200--499, 500+) to accommodate the heavy-tailed citation distribution (fewer papers at high citation levels). This design balances statistical precision and representativeness.

\section{Results}
\label{sec:results}

\subsection{Correlation Between LLM-Metrics and Citation Counts}

This section reports the core correlation analysis between LLM-Metrics and citation counts. We computed independent LLM-Metrics scores for each of the 17 models on 549 papers, then measured the Spearman rank correlation between each model's LLM-Metrics and raw citation counts to test whether model memory scores can predict research impact.

Overall, 15 of 17 models produced positive predictions, with 9 reaching statistical significance ($p < 0.05$) and 4 reaching high significance ($p < 0.001$). The overall LLM-Metrics (ensemble of all models) achieved a Spearman correlation of $\rho = 0.1495$ ($p = 0.0004$) and a Pearson correlation with log-transformed citations of $r = 0.1446$ ($p = 0.0007$). The statistical significance and cross-model directional consistency indicate that the memory signal is not random noise: it contains measurable information related to later citation impact.

This core finding carries multiple implications. First, the parametric memory of LLMs contains a real citation-related signal, consistent with the idea that model memory reflects exposure patterns in academic text. Second, 17 independently trained models produced directionally aligned predictions (15/17 positive), showing that the result cannot be reduced to the idiosyncratic bias of a single model. Third, the cross-model consistency of LLM-Metrics demonstrates that the metric is reproducible across models and vendors, although the magnitude of the signal remains model-dependent.

To further quantify the degree of cross-model consistency, we conducted two supplementary analyses. First, we computed the sign agreement rate: 15 out of 17 models (88.2\%) produced a positive Spearman $\rho$, significantly exceeding the chance expectation of 50\% (binomial test, $p = 0.0012$). This means that even under the most conservative criterion (distinguishing only positive versus negative direction, ignoring effect magnitude), model predictions are highly consistent. Second, we computed pairwise Spearman correlations between the LLM-Metrics scores produced by different models on the same set of papers as a measure of agreement. The mean pairwise correlation was $r = 0.44$ (range: 0.12--0.78), indicating moderate-to-strong agreement in paper-level memory assessments across independently trained models. Specifically, models from the same vendor tended to show higher pairwise agreement---the Meta LLaMA-3 family (4 models) had a mean pairwise $r = 0.62$, and the Alibaba Qwen2.5 family (7 models) had a mean pairwise $r = 0.50$. In contrast, cross-vendor model pairs had a mean pairwise $r = 0.40$. This pattern is explainable: models from the same vendor share more training data, data preprocessing pipelines, and architectural design choices, leading to more similar memory patterns; cross-vendor models diverge in both training data and architecture, naturally resulting in partially differentiated memory patterns.

Cross-model consistency is particularly noteworthy because it shows that LLM-Metrics is not purely model-specific. If each model remembered an entirely different set of papers due to idiosyncratic training data, LLM-Metrics would be difficult to use as a general metric. Instead, multiple independently trained models show measurable agreement in paper-level memory assessments. This agreement indicates that some papers leave more consistent traces across the training ecosystems of multiple LLMs because they are more widely represented in academic text sources. This convergence supports LLM-Metrics as a scalable scientometric indicator.

\begin{figure}[htbp]
\centering
\includegraphics[width=\textwidth]{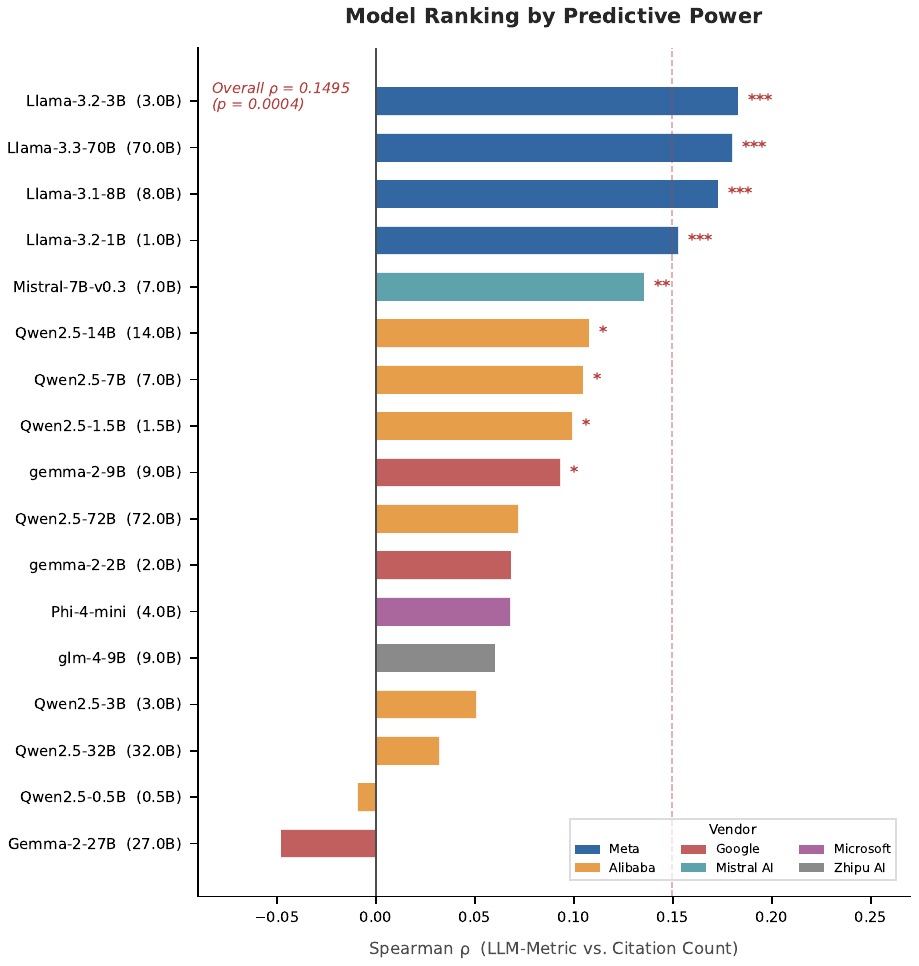}
\caption{\textbf{Model ranking by predictive power.} Horizontal bar chart showing Spearman $\rho$ between LLM-Metrics and citation counts for all 17 models, colored by vendor. Significance levels: $^{***}p<0.001$, $^{**}p<0.01$, $^{*}p<0.05$. The dashed vertical line marks the overall $\rho = 0.1495$. Model sizes (in billions of parameters) are shown in parentheses.}
\label{fig:ranking}
\end{figure}

Figure~\ref{fig:ranking} presents the complete ranking of all 17 models by Spearman correlation. The best-performing model is Llama-3.2-3B-Instruct ($\rho = 0.1829$, $p < 0.001$), followed closely by Llama-3.3-70B-Instruct ($\rho = 0.1801$, $p < 0.001$) and Llama-3.1-8B-Instruct ($\rho = 0.1729$, $p < 0.001$). Notably, the top four models all belong to the Meta LLaMA-3 family. A possible explanation is that Meta LLaMA-3 models have a higher relative capacity for open-access academic papers compared to other vendors (e.g., Alibaba, Google): LLaMA models are typically trained on a larger and more diverse web corpus, which may make them more sensitive to academic content memory. The bottom-ranked models---Gemma-2-27B ($\rho = -0.0480$, $p = 0.2614$) and Qwen2.5-0.5B-Instruct ($\rho = -0.0095$, $p = 0.8236$)---show virtually no predictive power, but their performance is consistent with the explanations from the extreme scale effect and capacity insufficiency aspects of the selective memory hypothesis (see \S4.5 for details).

\subsection{Monotonicity Across Citation Bins}

\begin{figure}[htbp]
\centering
\includegraphics[width=\textwidth]{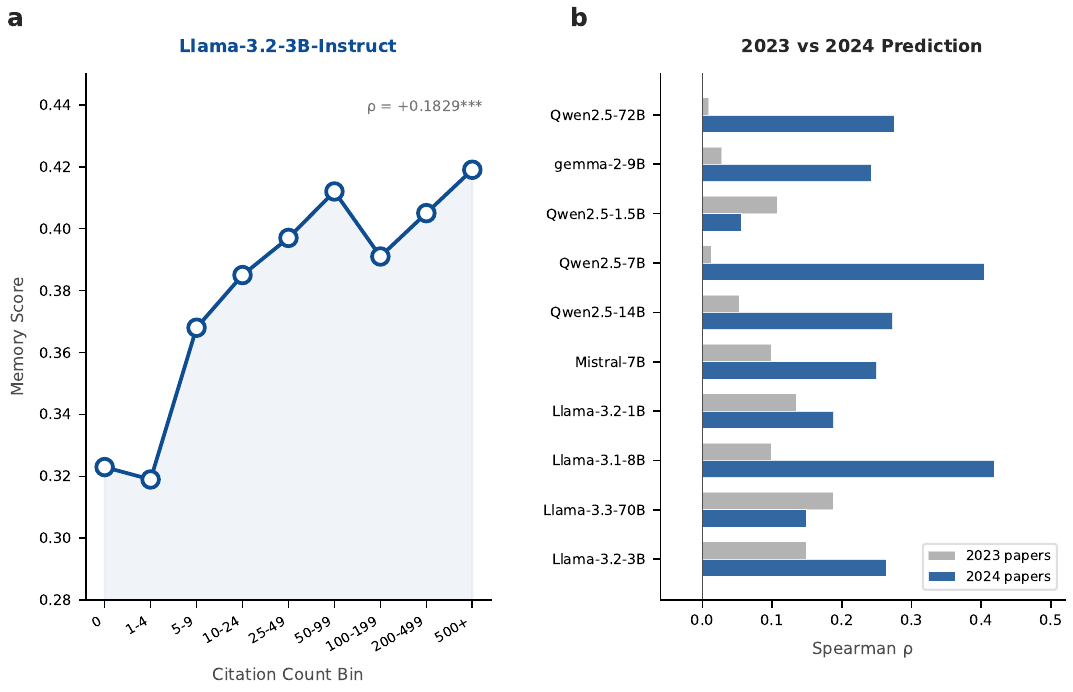}
\caption{\textbf{Mechanism evidence.} \textbf{a}, Memory scores across citation bins for Llama-3.2-3B-Instruct. The line plot shows the mean memory score for each citation bin, with the shaded region indicating the 95\% confidence band. Memory scores increase with citation counts ($\rho = +0.1829$, $p < 0.001$). \textbf{b}, Comparison of Spearman $\rho$ for 2023 versus 2024 papers across 10 models. The predictive signal is consistently stronger for 2024 papers (blue), which had near-zero citations at the time of model training, reducing the plausibility of a simple citation-count memorization explanation.}
\label{fig:mechanism}
\end{figure}

To further examine the structure of the signal, we selected the best-performing model, Llama-3.2-3B-Instruct, for a citation-bin monotonicity analysis. We divided papers into 9 citation bins (0, 1--4, 5--9, 10--24, 25--49, 50--99, 100--199, 200--499, 500+), computed the mean memory score for each bin, and tested whether memory scores increase with citation counts (Figure~\ref{fig:mechanism}a).

Results show a clear upward trend in memory scores with citation counts. From zero-citation papers (mean memory score 0.323) to highly cited papers with 500+ citations (mean memory score 0.419), the memory score increased by approximately 9.6 percentage points, corresponding to an effect size of approximately 0.30 standard deviations. Notably, in the low-citation range (0 to 1--4), the memory score slightly decreased (0.323 $\to$ 0.319) before resuming an upward trend starting from 5+ citations. This subtle V-shaped pattern may reflect the distinct nature of zero-citation papers: among zero-citation papers, some may be high-quality papers published very recently that have not yet accumulated citations but have already entered the model's training data through other exposure channels (e.g., social media discussions, arXiv preprint dissemination), yielding a detectable memory signal. From the 5--9 bin onward, memory scores rise steadily and plateau at the 500+ bin (0.419).

More importantly, all three component dimensions of the memory score---CORRECT rate, WRONG rate, and REFUSAL rate---exhibited directions consistent with theoretical expectations. Highly cited papers had higher CORRECT rates ($\rho = +0.1829$, $p < 0.001$), meaning the model was more likely to correctly answer probes about high-impact papers. Simultaneously, highly cited papers had lower WRONG rates ($\rho = -0.1230$, $p = 0.004$) and lower REFUSAL rates ($\rho = -0.1081$, $p = 0.011$), indicating that when facing high-impact papers, the model not only answered correctly more often but also answered incorrectly or refused to answer less often. The directional consistency across all three dimensions constitutes a ``triple cross-validation'': positive signals (higher CORRECT) and negative signals (lower WRONG and REFUSAL) jointly point to the same conclusion, substantially reducing the likelihood that the signal is a statistical artifact.

Figure~\ref{fig:mechanism}a visually illustrates this trend: the CORRECT rate increased by approximately 9.6 percentage points from the low-citation bin (0 citations: 32.3\%) to the high-citation bin (500+: 41.9\%); the WRONG rate decreased by approximately 5.3 percentage points from the low-citation bin (0 citations: 33.4\%) to the high-citation bin (500+: 28.1\%); and the REFUSAL rate reached its minimum in the mid-citation range (50--99: 28.9\%) and was slightly higher at both extremes. This pattern suggests that models behave differently toward papers at the extremes of the impact distribution: for low-impact papers, models refuse more often due to lack of knowledge; for high-impact papers, models answer correctly more often due to having the knowledge; mid-impact papers fall in a transitional zone.

This monotonicity analysis supports the construct validity of LLM-Metrics as a research-impact memory signal. If LLM-Metrics measures impact-related exposure, it should increase along the impact gradient; the observed trend is consistent with this expectation. Together with the CORRECT/WRONG/REFUSAL decomposition, the citation-bin analysis provides direct evidence that model memory is stronger for higher-impact papers.

\subsection{Temporal Split: A Clean Test from 2024 Papers}

A key challenge for the selective-memory hypothesis is the reverse-causality problem: the correlation between LLM-Metrics and citation counts could arise because models ``saw'' citation information in their training data (e.g., citation counts appearing on academic search pages or paper header annotations), rather than because models remembered paper content or surrounding academic discussion. If this were the case, LLM-Metrics would be merely an indirect citation-capture mechanism, offering limited information beyond citation counts themselves.

To test this alternative explanation, we leveraged a natural temporal contrast between model training cutoffs and paper publication dates. Our sample contains 415 papers published in 2023 and 134 papers published in 2024. Model training cutoffs are concentrated between May and December 2024. Papers published in 2024 had only 0--6 months of citation window before model training cutoffs, with most having 0--2 citations. If LLM-Metrics only captures citation-count information, its predictive power should weaken for 2024 papers. Conversely, if LLM-Metrics captures broader exposure traces---for example, arXiv full texts, academic discussions, project pages, blogs, or news dissemination---it can remain predictive even when mature citation counts are unavailable.

\begin{table}[htbp]
\centering
\caption{LLM-Metrics correlation by publication year.}
\label{tab:temporal}
\begin{tabular}{lcc}
\toprule
\textbf{Year} & \textbf{Papers} & \textbf{Overall $\rho$} \\
\midrule
2023 & 415 & +0.0559 \\
2024 & 134 & +0.1880 \\
\bottomrule
\end{tabular}
\end{table}

The results support the latter prediction. The overall Spearman $\rho$ for 2024 papers was 0.1880---more than three times that of 2023 papers ($\rho = 0.0559$) (Table~\ref{tab:temporal}). Among the 10 models with valid scores for both cohorts, 8 showed higher $\rho$ for 2024 than for 2023 (Figure~\ref{fig:mechanism}b). Because these papers had near-zero citations at model-training time, this pattern rules out a simple reverse-causality explanation in which LLM-Metrics merely captures mature citation counts.

Individual model performances illustrate the same pattern. Qwen2.5-7B-Instruct achieved $\rho = 0.4044$ for 2024 papers versus only 0.0124 for 2023. Similarly, Llama-3.1-8B-Instruct achieved $\rho = 0.4181$ for 2024 papers versus 0.0986 for 2023, and Qwen2.5-72B-Instruct achieved $\rho = 0.2754$ for 2024 papers versus 0.0088 for 2023. These contrasts suggest that recent papers can carry a stronger memory-based signal before citation counts mature. Notably, Llama-3.3-70B-Instruct was one of the few models whose 2023 $\rho$ (0.1873) exceeded its 2024 $\rho$ (0.1486), possibly because its later training cutoff (December 2024) exposed it to more mature metadata for some 2024 papers.

This temporal split provides one of the cleanest pieces of evidence for the selective-memory hypothesis. The most parsimonious interpretation is that LLM-Metrics captures exposure traces in training corpora---including full-text content, blog discussions, news dissemination, and other channels---rather than mature citation counts. We therefore interpret the memory-citation association as the downstream footprint of a shared impact process: paper quality, author visibility, topic salience, and community attention jointly influence both academic exposure and future citations, and LLM-Metrics measures the memory trace left by this exposure.

\subsection{Differential Discriminative Power Across Probe Types}

LLM-Metrics is composed of four distinct probe types---title recognition, author recognition, method recognition, and venue information recognition. If the selective-memory hypothesis holds, different probe types should exhibit different discriminative power: probes most directly related to academic exposure frequency should show the strongest power, while probes relying on generalizable knowledge or universally accessible information should be weaker. This section systematically compares the discriminative power across probe types (Table~\ref{tab:probe_power}).

\begin{table}[htbp]
\centering
\caption{Discriminative power across probe types.}
\label{tab:probe_power}
\begin{tabular}{lcc}
\toprule
\textbf{Probe Type} & \textbf{Power} & \textbf{Mechanistic Interpretation} \\
\midrule
E2\_mc (Author Recognition) & Strongest & Papers by prominent authors receive greater exposure \\
E1\_mc (Title Recognition) & Moderate & Title memory linked to paper exposure frequency \\
E4\_mc (Method-Paper) & Weakest & Method knowledge is generalizable \\
F1\_mc (Venue Information) & Partial & Venue information equally accessible for all papers \\
\bottomrule
\end{tabular}
\end{table}

Author recognition probes (E2\_mc) showed the strongest discriminative power among all four types. This result is consistent with the Matthew effect in academia: papers by prominent authors can simultaneously attract higher citation counts and greater training-data exposure. In academic communication, author identity is a salient dimension of paper exposure; new papers by well-known authors are more likely to be discussed, shared, and cited. Model training data may reflect this exposure disparity, leading to stronger encoding of author information. This finding supports the proposed mechanism, while also highlighting a fairness caveat: LLM-Metrics may partially encode author visibility rather than paper merit alone.

Title recognition probes (E1\_mc) showed the second-strongest discriminative power. Paper titles are among the most distinctive identifiers of a paper, and title memory directly reflects the degree to which the model has encoded the paper itself. High-impact papers, due to more frequent discussion and citation, have their titles appearing more frequently in training data, leading to more robust encoding in model parameters. As the most natural and most numerous probe type (2,745 probes), the moderate discriminative power of title probes may be because title information is relatively brief (typically 10--20 words), and models can encounter the same title fragments across different contexts in training data, to some extent blurring memory differences between high-impact and low-impact papers.

Method-paper association probes (E4\_mc) showed the weakest discriminative power. This result aligns with theoretical expectations: methodological knowledge is generalizable---for example, the knowledge that ``ResNet is used for image classification'' can be learned by models from diverse sources (the original paper, subsequent citing papers, tutorials, code documentation, etc.) without depending on memory of any specific paper. Consequently, a model's correct answer to a method probe does not necessarily reflect memory of a particular paper but may reflect learning from a broader knowledge distribution. This reduces the effectiveness of method-based probes for paper-level impact discrimination.

Venue information probes (F1\_mc) showed discriminative power only in a subset of models. One possible reason is that venue and year information is equally accessible for all papers---whether high-impact or low-impact, venue and year can be extracted from PDF metadata or citation pages. Therefore, the frequency of venue information appearing in training data may be relatively uniform across papers, failing to constitute an effective discrimination signal. However, the partial discriminative power observed in some models suggests that certain models' training data may contain venue information exposure disparities---for example, abstract pages from specific conferences may be more completely ingested.

These differentiated results not only validate the selective-memory hypothesis but also carry practical methodological implications: in future applications of LLM-Metrics, one can select or weight different probe types based on evaluation objectives. For maximizing discriminative power, a combination of author recognition and title recognition probes can be prioritized; for a more balanced multi-dimensional assessment, the integration of all four probe types can be employed.

\subsection{Non-Monotonic Scaling with Model Size}

\begin{figure}[htbp]
\centering
\includegraphics[width=\textwidth]{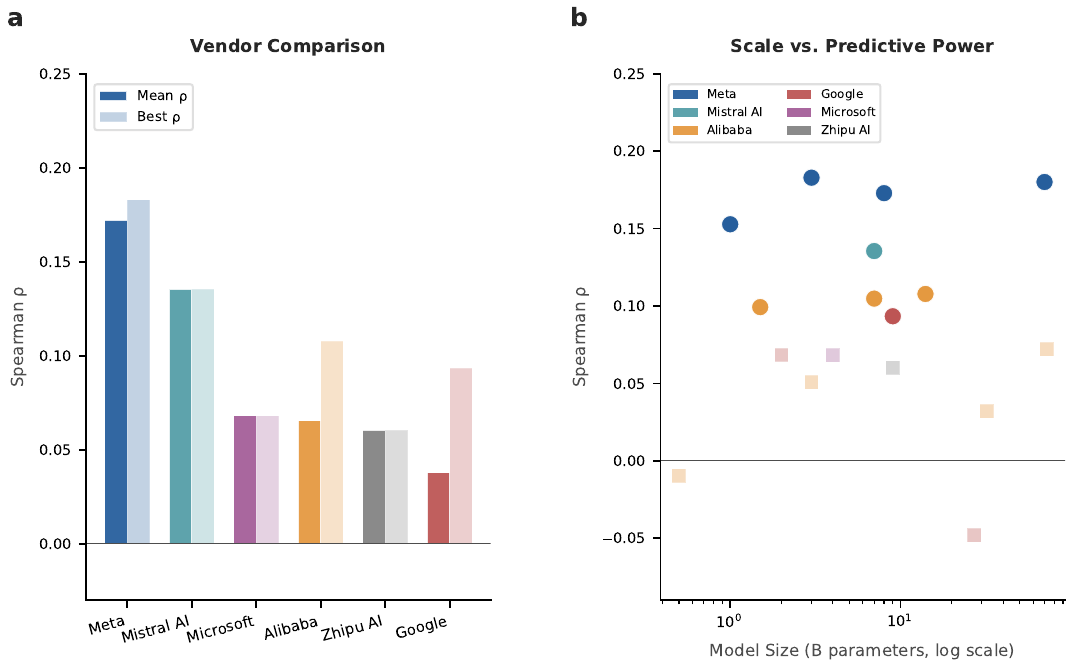}
\caption{\textbf{Vendor comparison and non-monotonic scaling.} \textbf{a}, Vendor-level comparison of mean and best Spearman $\rho$. Meta's LLaMA-3 family consistently outperforms other vendors. \textbf{b}, Scatter plot of model size versus Spearman $\rho$, colored by vendor. Filled circles indicate significant correlations ($p < 0.05$); open squares indicate non-significant. The relationship is non-monotonic: a 3B-parameter model (Llama-3.2-3B-Instruct) achieves the highest $\rho$, outperforming most larger models.}
\label{fig:vendor_scaling}
\end{figure}

If LLM-Metrics merely reflected a ``larger models remember more, therefore predict better'' pattern, predictive power should increase monotonically with parameter count. The data, however, contradict this intuitive prediction and reveal a more complex, theoretically informative relationship (Figure~\ref{fig:vendor_scaling}b).

Examining average predictive power by size group: the 0--4B group (6 models) averaged $\rho = 0.0908$; the 4--10B group (6 models) averaged $\rho = 0.1059$; the 10--30B group (2 models) averaged $\rho = 0.0299$; the 30--100B group (3 models) averaged $\rho = 0.0948$. The highest average predictive power was not in the largest-parameter group but in the intermediate group (4--10B), and the lowest was in the 10--30B group rather than the smallest. At the individual model level, the top-ranked predictor was Llama-3.2-3B-Instruct (3.0B parameters, $\rho = 0.1829$), outperforming all 7B, 8B, 14B, 32B, and most 70B+ parameter models. One of the worst predictors was Gemma-2-27B (27.0B parameters, $\rho = -0.0480$).

This counterintuitive non-monotonic pattern is consistent with the selective-memory hypothesis. We interpret it as a possible ``selective-memory window'' effect. Very small models (e.g., 0.5B--1.5B) have limited knowledge capacity and may memorize too few papers to provide stable discrimination (e.g., Qwen2.5-0.5B-Instruct, $\rho = -0.0095$). As model size grows into the 3B--8B range, models may acquire enough capacity to encode many academic papers, yet still not enough to encode the long tail uniformly. This creates a potential \textbf{discrimination window}: high-exposure papers are preferentially encoded, whereas low-exposure papers remain weakly represented. When model size further increases, training data composition, alignment, deduplication, and model behavior may dominate simple capacity effects. The reversal of Gemma-2-27B is therefore best treated as evidence that scale alone is insufficient, rather than as proof of a universal ceiling effect.

This finding has two implications. Theoretically, the relationship between model memory and model scale is not a simple linear function. Practically, LLM-Metrics should be calibrated on held-out validation data before use; researchers should not assume that the largest available model is the most informative model for memory-based scientometric analysis.

\subsection{Vendor-Level Comparison}

Models from different vendors showed marked and systematic differences in LLM-Metrics' predictive performance (Figure~\ref{fig:vendor_scaling}a), and these differences contain important information about the relationship between training data composition and model evaluation capability.

Among the six vendors, Meta's LLaMA-3 family showed the strongest performance, with all four models achieving an average Spearman $\rho = 0.1722$ and all reaching high statistical significance at $p < 0.001$. More notably, these four models span a large parameter range (1.0B to 70.0B), yet their predictive power varies little ($\rho$ range: 0.1528--0.1829). One plausible explanation is that the LLaMA-series training mixture contains enough academic and knowledge-domain text to support stable paper memory. Because exact training mixtures are not fully observable, we treat this as a hypothesis rather than a definitive attribution.

Mistral AI's Mistral-7B-Instruct-v0.3 (single model) ranked second ($\rho = 0.1355$, $p = 0.0015$), but with only one model from this vendor evaluated, it is difficult to assess whether its performance is robust across scales.

Alibaba's Qwen2.5 family had the largest number of models (7), yet its average predictive power was relatively low ($\rho = 0.0654$) with enormous internal variation: the best was Qwen2.5-14B-Instruct ($\rho = 0.1078$), and the worst was Qwen2.5-0.5B-Instruct ($\rho = -0.0095$), the only Qwen2.5 model producing a negative correlation. The high internal variance may reflect diversity in training data mixtures, training strategies, or model initialization within the Qwen2.5 family. Notably, two models in the Qwen2.5 family (3B and 32B) had refusal rates exceeding 97\% (correct rate only 1--2\%), which severely limits their discriminability because when a model refuses the vast majority of questions, it cannot provide an effective discrimination signal. These models may be more suitable for high-precision confirmation tasks rather than large-scale screening.

Google's Gemma-2 family exhibited the most anomalous pattern: performance improved from 2B to 9B but reversed sharply at 27B---gemma-2-2b-it ($\rho = 0.0684$) $<$ gemma-2-9b-it ($\rho = 0.0934$) $>$ Gemma-2-27B ($\rho = -0.0480$). The 27B-parameter model was the only one in the family to produce a negative correlation, and its correct rate (17.7\%) was substantially lower than those of the 2B (41.8\%) and 9B (41.0\%) models. This scale reversal may reflect differences in training data, distillation, alignment, or response style; the present experiments cannot isolate which factor is responsible.

Microsoft's Phi-4-mini-instruct and Zhipu AI's glm-4-9b-chat each had one model evaluated, achieving $\rho = 0.0682$ and $\rho = 0.0603$, respectively, neither reaching statistical significance ($p = 0.1103$ and $p = 0.1584$), with predictive power in the low-to-moderate range. These results suggest that paper memory discriminability may not be a universal property of all LLMs but is specific to those whose training data contain a sufficiently large and diverse volume of academic corpora.

In summary, vendor-level differences reveal a core practical point: the usefulness of LLM-Metrics depends on the evaluated model, and likely on its training data composition and response behavior, rather than merely on model scale or vendor reputation. Choosing and validating the model is therefore part of the measurement procedure.

\section{Discussion}
\label{sec:discussion}

\subsection{LLM-Metrics as a New Scientometric Indicator}

Our results establish LLM-Metrics as a new memory-based scientometric indicator. Across 17 independently trained models, it produces a reproducible signal related to later citation impact. Compared with traditional citation-based metrics, LLM-Metrics has four distinctive advantages. First, it can be computed without waiting for multi-year citation accumulation. Second, it integrates multiple memory dimensions (title, author, method, and venue memory). Third, it is naturally extensible across fields because it only requires paper metadata and model queries. Fourth, it is updatable: as new models are released, LLM-Metrics can be recomputed to study how model-mediated scholarly memory changes over time.

\subsection{Validation of the Selective-Memory Hypothesis}

We evaluated the selective-memory hypothesis through four converging lines of evidence. The citation-bin analysis showed that memory scores increase with citation counts in aggregate. The temporal split test revealed stronger predictive signals for 2024 papers, which had near-zero citations at training time, reducing the plausibility of a simple citation-count memorization explanation. The probe-type analysis found that author recognition, a probe type closely linked to academic visibility, had the strongest discriminative power. Finally, the non-monotonic scaling analysis showed that model scale is not sufficient to explain predictive power.

\subsection{Vendor-Level Differences}

The substantial variation in predictive performance across vendors has practical implications. Meta's LLaMA-3 family consistently outperformed other vendors, possibly reflecting a higher proportion of academic content in its training data. Alibaba's Qwen2.5 family showed high internal variance, suggesting that model-level factors beyond vendor identity influence predictive power. Google's Gemma-2 family exhibited a reversal at scale, with the largest model (27B) producing a negative correlation, indicating that training data composition may be more important than model scale for LLM-Metrics' effectiveness.

The cross-model consistency analysis provides additional evidence for LLM-Metrics' robustness. The 88.2\% sign agreement rate across 17 independently trained models and the moderate-to-strong pairwise correlations (mean $r = 0.44$) between model-level LLM-Metrics scores indicate that the memory signal is unlikely to be an artifact of a single model. Rather, it reflects a repeated pattern in how academic papers are differentially represented across multiple LLMs. This convergence strengthens the case for LLM-Metrics as a scientometric metric driven by model memory.

\subsection{Comparison with Traditional Citation Prediction Methods}

LLM-Metrics differs from existing citation prediction approaches in what it measures. Traditional methods (e.g., ForeCite~\citep{hull2025forecite}, CiMaTe~\citep{hirako2024cimate}) rely on textual features, author networks, or early citation patterns to predict future citation counts, essentially modeling ``what features characterize highly cited papers.'' LLM-Metrics instead measures the model's ``memory strength'' of papers and asks whether that memory strength is informative about later citation impact. This distinction is methodologically useful: LLM-Metrics requires no task-specific training or feature engineering, but it also inherits the opacity and bias of the underlying model's training data.

In terms of raw performance, LLM-Metrics' overall Spearman $\rho = 0.1495$ is lower than that of dedicated citation prediction models (e.g., ForeCite achieves $\rho = 0.826$ on biomedical papers). This difference is expected: dedicated models are trained on large-scale, domain-specific labeled data, whereas LLM-Metrics is zero-shot---requiring no task-specific training, no citation inputs, and no feature engineering. Its value lies in a different regime: LLM-Metrics provides a citation-independent exposure-memory metric that can be computed immediately once a suitable model is available and can be combined with text, metadata, early citations, and altmetrics in future systems.

\subsection{Scope and Limitations}

Several scope conditions should be acknowledged. First, our sample contains 549 papers and should be expanded in future arXiv-scale and cross-disciplinary studies. Second, model training corpora are not fully observable; LLM-Metrics measures the memory trace left by exposure even when the exact source of that exposure cannot be identified. Third, model selection is constrained to publicly available models with approximate 2024 training cutoffs; earlier or later models may exhibit different patterns. Fourth, author prestige, institutional visibility, topic popularity, and social media exposure may all influence both training-data frequency and citation accumulation; these factors are part of the scholarly exposure process that LLM-Metrics is designed to capture, but they also require fairness-aware interpretation. Fifth, the two-to-three-year post-publication window is relatively short; citation trajectories may shift over longer periods. Sixth, our probe design is limited to multiple-choice format; open-ended probes may capture different aspects of model memory.

\subsection{Future Directions}

Several promising directions emerge. First, extending the framework to other disciplines (biology, physics, social sciences) would test the cross-disciplinary generalizability of LLM-Metrics. Different disciplines exhibit vastly different scholarly communication patterns---biology relies on preprints (bioRxiv), physics has long used arXiv, and social sciences depend more on journals and monographs---and these differences may cause LLM-Metrics to exhibit distinct predictive characteristics across fields. Second, tracking how LLM-Metrics changes across successive model generations could provide a dynamic measure of model-mediated scholarly memory. Third, exploring more sophisticated probing techniques, including open-ended recall, calibrated confidence estimation, and adversarial distractor construction, may improve recall and discriminative power. Fourth, investigating the relationship between training data composition and predictive performance could clarify which sources (e.g., arXiv full texts, academic blogs, conference websites) drive the signal. Fifth, future work should test whether LLM-Metrics adds incremental predictive value beyond text embeddings, metadata, early citations, author features, and altmetrics in a unified forecasting model.

\section{Conclusion}

We have proposed LLM-Metrics, a paper-level research-impact metric grounded in the parametric memory of large language models, and evaluated it across 17 LLMs spanning six vendors and 0.5B--72B parameters on 549 computer science papers. The findings are fivefold. First, LLM-Metrics contains a statistically reliable citation-related signal: 15 of 17 models produced positive predictions (9 significant), with an overall $\rho = 0.1495$ ($p = 0.0004$). Second, the selective-memory hypothesis is supported by four converging lines of evidence: citation-bin monotonicity, the temporal split test, probe-type differentiation, and non-monotonic scaling. Third, vendor-level differences are substantial: Meta's LLaMA-3 family consistently outperforms others, while Google's Gemma-2 family exhibits a reversal at scale. Fourth, model scale and predictive power are non-monotonic: a 3B-parameter model outperforms most larger models. Fifth, cross-model consistency analysis (88.2\% sign agreement rate, mean pairwise $r = 0.44$) supports the reproducibility of the memory signal.

LLM-Metrics offers a citation-independent way to assess research impact by measuring how scholarly exposure is reflected in large language model memory. Its core contribution is to treat LLMs not only as objects of evaluation, but also as instruments for measuring traces of academic dissemination. The resulting signal is reproducible across many independently trained models and is available without task-specific training. As LLMs become increasingly embedded in the academic ecosystem, memory-based indicators such as LLM-Metrics provide a real-time, cross-disciplinary, citation-independent paradigm for research assessment.

\bibliographystyle{plainnat}
\bibliography{citations/references}

\end{document}